\documentclass[letterpaper, 10 pt, conference]{ieeeconf}  
 \IEEEoverridecommandlockouts 
\usepackage{amsfonts,amssymb,amsmath,epsfig}
\usepackage{fullpage}
\usepackage{color}
\usepackage{array}
\usepackage{hyperref}
\usepackage{fancyhdr}

\hypersetup{
    colorlinks=true, 
    linkcolor=black,  
    citecolor=black,    
    filecolor=black,     
    urlcolor=black     
}

\newcommand{\be}{\begin{equation}}
\newcommand{\ee}{\end{equation}}
\newcommand{\ba}{\left[ \begin{array}}
\newcommand{\ea}{\end{array} \right]}
\newcommand{\bea}{\begin{eqnarray}}
\newcommand{\eea}{\end{eqnarray}}

\def\real{\mathbb{R}}

\def\x{{\bf{x}}}
\def\y{{\bf{y}}}
\def\n{{\bf{n}}}
\def\w{\omega}

\def\1{^{\prime}}

\def\ignore#1{}
\def\imu{_\mathrm{imu}}

\def\V{{v}}

\begin{document}

\title{\bf Robust Inference for Visual-Inertial Sensor Fusion}
\author{Konstantine Tsotsos$^1$ \and Alessandro Chiuso$^2$ \and Stefano Soatto$^1$} \date{}
\maketitle
\thispagestyle{fancy}

\begin{abstract}
Inference of three-dimensional motion from the fusion of inertial and visual sensory data has to contend with the preponderance of outliers in the latter. Robust filtering deals with the joint inference and classification task of selecting which data fits the model, and estimating its state. We derive the optimal discriminant and propose several approximations, some used in the literature, others new. We compare them analytically, by pointing to the assumptions underlying their approximations, and empirically. We show that the best performing method improves the performance of state-of-the-art visual-inertial sensor fusion systems, while retaining the same computational complexity. \\

\raggedright Supplementary video results available at: \href{http://youtu.be/5JSF0-DbIRc}{http://youtu.be/5JSF0-DbIRc}
\end{abstract}

\section{Introduction}

\footnotetext[1]{K. Tsotsos and S. Soatto are at the University of California, Los Angeles, USA. Email: \href{mailto:ktsotsos@cs.ucla.edu}{\{ktsotsos,soatto\}@cs.ucla.edu}}
\footnotetext[2]{A. Chiuso is at the University of Padova, Italy. Email: \href{mailto:chiuso@dei.unipd.it}{chiuso@dei.unipd.it}}

Low-level processing of visual data for the purpose of three-dimensional (3D) motion estimation yields mostly garbage: Easily $60-90\%$ of sparse features selected and tracked across frames are inconsistent with a single rigid motion due to illumination effects, occlusions, or independently moving objects. These effects are global to the scene, while low-level processing is local to the image, so it is not realistic to expect significant improvements in the vision front-end. Instead, it is paramount that inference algorithms that use vision be capable of dealing with such a preponderance of ``outlier'' measurements. This includes leveraging on other sensory modalities, such as inertials. We tackle the problem of inferring ego-motion of a sensor platform from visual and inertial measurements, focusing on the handling of outliers. This is a particular instance of robust filtering, a mature area of statistics, and most visual-inertial integration systems (VINS) employ some form of inlier/outlier test. Different VINS use different methods, making their comparison difficult. None relate their approach analytically to the optimal (Bayesian) classifier.  

We derive the optimal discriminant, which is intractable, and describe different approximations, some currently used in the VINS literature, others new. We compare them analytically, by pointing to the assumptions underlying their approximations, and empirically. The results show that it is possible to improve the performance of a state-of-the-art system with the same computational footprint.
\subsection{Related work}
\vspace{-0.2cm}
The term ``robust'' in filtering and identification refers to the use of inference criteria that are more forgiving than the ${\mathbb L}^2$ norm. They can be considered special cases of Huber functions \cite{huber81}, where the residual is re-weighted, rather than data selected (or rejected). More importantly, the inlier/outlier decision is typically {\em instantaneous}.
Our derivation of the optimal discriminant follows from standard hypothesis
testing (Neyman-Pearson), and motivates the introduction of a delay-line in the
model, and correspondingly the use of a ``smoother'', instead of a standard
filter. State augmentation with a delay-line is common practice in the design
and implementation of observers and controllers for so-called ``time-delay
systems'' \cite{trinh1997memoryless,bhat1976observer} or ``time lag systems''
\cite{leyva1995asymptotic,rao1979identification} and has been used in VINS
\cite{eustice2004visually,roumeliotis2002augmenting}.
Various robust inference solutions proposed in the navigation and SLAM literature
(simultaneous localization and mapping), such as One-point Ransac
\cite{civera20121}, or MSCKF \cite{mourikisR07}, can also be related to the
standard approach. Similarly, \cite{neira2001data} maintains a temporal window to re-consider inlier/outlier
associations in the past, even though it does not maintain an estimate of the
past state.

\subsection{Notation and mechanization}

We adopt the notation of \cite{murrayLS94,maSKS}:
The {\em spatial frame} $s$ is attached to Earth and oriented so gravity
$\gamma^T = [0 \ 0 \ 1]^T \| \gamma \|$ is known.
The {\em body frame} $b$ is attached to the IMU. The {\em camera frame} $c$ is also unknown, although
{\em intrinsic calibration} has been performed, so that measurements are in metric units. 
The equations of motion (``mechanization'') are described in the body frame at time $t$ relative to the spatial frame $g_{sb}(t)$. Since the spatial frame is arbitrary, it is co-located with the body at $t = 0$. To simplify the notation, we indicate $g_{sb}(t)$ simply as $g$, and so for $R_{sb}, T_{sb}, \w_{sb}, v_{sb}$, thus omitting the subscript $sb$ wherever it appears. This yields a model for pose $(R,T)$, linear velocity $v$ of the body relative to the spatial frame:
\begin{equation}
{
\begin{cases}
\begin{tabular}{>{$}r<{$} >{$\!\!\!\!\!}l<{$} >{$}r<{$} >{$\!\!\!\!\!}l<{$}}
\dot T &= \V &T(0) &= 0 \\
\dot R &= R (\widehat \w\imu  - \widehat \w_b) + n_{R} &R(0) &= R_0\\
\dot \V &= R(\alpha\imu  - \alpha_b) + \gamma + n_\V \\ 
\dot \w_b &= w_b \\
\dot \alpha_b &= \xi_b. 
\end{tabular}
\end{cases}
}
\label{eq-standard}
\end{equation}
where gravity $\gamma \in \real^3$ is treated as a known parameter, $\w\imu$ are
the gyro measurements, $\w_b$ their  unknown bias, $\alpha\imu$ the accel
measurements and $\alpha_b$ their unknown bias.

Initially we assume there is a collection of points $p_i$ with coordinates $X_i
\in \real^3, \ i = 1, \dots, N$, visible from time $t=0$ to the current time
$t$. If $\pi:\real^3 \rightarrow \real^2; X \mapsto [X_1/X_3, \ X_2/X_3]$ is a
canonical central (perspective) projection, assuming that the camera is
calibrated and that the spatial frame coincides with the body frame at time $0$,
a point feature detector and tracker \cite{lucasK81a} yields $y_i(t)$, for all
$i = 1, \dots, N$, 
\begin{equation}
{y_i(t) =  \pi(g^{-1}(t)p_i) + n_i(t), ~~~ {\color{black} t \ge 0}}
\label{eq-y}
\end{equation}
where $\pi(g^{-1}(t)p_i)$ is represented in coordinates as ${ \frac{R^T_{1:2}(t) (X_i - T(t))}{R^T_{3}(t)( X_i - T(t))}}$. 
In practice, the measurements $y(t)$ are known only up to an ``alignment'' $g_{cb}$  mapping the body frame to the camera: 
\begin{equation}
{y_i(t) = \pi\left( g_{cb} g^{-1}(t) p_i \right) + n_i(t) \in \real^2}
\label{eq-vis}
\end{equation}
The unknown (constant) parameters $p_i$ and $g_{cb}$ can then be added to the state with trivial dynamics: 
\begin{equation}
\begin{cases}
\begin{tabular}{>{$}r<{$} >{$\!\!\!\!\!}l<{$}}
\dot p_i &= 0, ~~~ i = 1, \dots, N(j) \\
\dot g_{cb} &= 0.
\end{tabular}
\end{cases}
\label{eq-groups}
\end{equation}
The model \eqref{eq-standard},\eqref{eq-groups} with measurements \eqref{eq-vis} can be written compactly by defining the state $x = \{T, R, v, \w_b, \alpha_b, T_{cb}, R_{cb} \} \doteq \{x_1, x_2, \dots, x_7 \}$ where $g_{cb} = (R_{cb}, T_{cb})$, and the structure parameters $p_i$ are represented in coordinates by $X_i = {{\bar y}_i}({t_i})\exp(\rho_i)$, which ensures that $Z_i = \exp(\rho_i)$ is positive. We also define the {\em known input} $u = \{\widehat\omega\imu , \alpha\imu \} = \{u_1, u_2\}$, the {\em unknown input} $v = \{ w_b, \xi_b \} = \{v_1, v_2\}$ and the model error $w = \{n_R, n_v\}$. After defining suitable functions $f(x)$, $c(x)$, matrix $D$  
and $h(x,p) = [\dots, \pi(x_2^T(p_i - x_1))^T, \dots]^T$
with $p =\{p_1, \dots, p_N\}$ the model \eqref{eq-standard},\eqref{eq-groups}, \eqref{eq-vis} takes the form 
\begin{equation}
{\begin{cases}
\dot x = f(x) + c(x) u + D v + c(x) w\\
\dot p = 0 \\ 
y = h(x,p) + n.
\end{cases}}
\label{eq-fg}
\end{equation}
To enable a smoothed estimate
 we augment the state with a delay-line: Let $g(t) \doteq (R(t), T(t))$. Then,
for a fixed interval $dt$ and $1 \le n \le k$,  define $x_{n}(t) \doteq g(t - n
dt)$, $\x^{k} \doteq \{x_{1}, \dots, x_{k}\}$ that satisfies
\be
\x^k(t+dt) 
\doteq F \x^k(t) + G  x(t)
\ee
where
\be
F \doteq \ba{cccc}
0 & & & \\
I & 0 &  & \\
  &   & \ddots & \\
0 & \ldots & I & 0
\ea, \  { G \doteq \ba{cccc} I  & 0  & \hdots & 0 \\ 0 & 0 & \hdots & 0 \\ \vdots & \vdots & \ddots & \vdots \\ 0 & 0 &\dots & 0\ea} 
\ee
and $\x \doteq \{x, x_1, \dots, x_{k}\} = \{x, \x^{k}\}$. A $k$-stack of measurements ${\y}_j^k(t) = \{y_j(t), y_j(t-dt), \dots, y_j(t - k dt)\}$ can be related to the smoother's state $\x(t)$ by
\be
{\y_j}(t) = h^k(\x(t), p_j) + {\n_j}(t) 
\ee
where we omit the superscript $k$ from $\y$ and $\n$, and 
\be
h^k(\x(t), p_j) \doteq \ba{c} h(x(t),  p_j)\ \pi(x_{1}(t) p_j) \dots
\pi(x_{k}(t) p_j) \ea^T \ee

Note that $\n_j$ is {\em not} temporally white even if $n_j$ is. The overall model is then
\be
\begin{cases}
\dot x = f(x) + c(x) u + D v + c(x) w \\
\x^k(t+dt) = F \x^k(t) + G x(t) \\
\dot p_j = 0 \\
\y_j(t) = h^k(\x(t),p_j) + {\n}_j(t), \\ ~~~~~~~~~~~~~~  t \ge t_j, \ j = 1, \dots, N(t)
\end{cases}
\label{eq-smoothing}
\ee
The observability properties of \eqref{eq-smoothing}, are the same as
\eqref{eq-fg}, and are studied in \cite{hernandezS13}, 
where it is shown that \eqref{eq-fg} is {\em not} unknown-input observable
(Claim 2), although it is observable with no unknown inputs \cite{jonesS09}.
This means that, as long as gyro and accel bias rates are not identically zero,
convergence of {\em any} inference algorithm to a unique point estimate cannot
be guaranteed. Instead, \cite{hernandezS13} explicitly computes the
indistinguishable set (Claim 1) and bounds it as
a function of the bound on the accel and gyro bias rates.

\section{Robust Filtering} 
\label{sect-filtering}

In addition to the inability of guaranteeing convergence to a unique point estimate, the major challenge of VINS is that the majority of imaging data $y_i(t)$ does not fit \eqref{eq-fg} due to specularity, transparency, translucency, inter-reflections, occlusions, aperture effects, non-rigidity and multiple moving objects. While filters that approximate the entire posterior, such as particle filters, in theory address this issue, in practice the high dimensionality of the state space makes them intractable. Our goal thus is to couple the inference of the state with a classification to detect which data are inliers and which are outliers, and discount or eliminate the latter from the inference process.

In this section we derive the optimal classifier for outlier detection, which is also intractable, and describe approximations, showing explicitly under what conditions each is valid, and therefore allowing comparison of existing schemes, in addition to suggesting improved outlier rejection procedures.  For simplicity, we assume that all points appear at time $t = 0$,  and are present at time $t$, so we indicate the ``history'' of the measurements {\em up to time $t$} as $y^t  = \{y(0), \dots, y(t)\}$ (we will lift this assumption in Sect. \ref{sect-implementation}).  We indicate inliers with $p_j$, $j \in J$, with $J \subset [1, \dots, N]$ the inlier set, and assume $| J | \ll N$, where $| J |$ is the cardinality of $J$. 

While a variety of robust statistical inference schemes have been developed for filtering \cite{benveniste,el2001robust,huber81,bar-shalomL98}, most operate under the assumption that the majority of data points are inliers, which is not the case here.

\subsection{Optimal discriminant}

In this section and the two that follow we will assume\footnote{The first assumption carries no consequence in the design of the discriminant, the latter will be lifted in Sect. \ref{sect-structure}.} that the inputs $u,v$ are absent and the parameters $p_i$ are known, which reduces \eqref{eq-fg} to the standard form 
\be
\begin{cases}
\dot x = f(x) + w \cr
y = h(x) + n.
\label{eq-simpl}
\end{cases}
\ee
To determine whether a datum $y_i$ is inlier, we consider the event ${\cal I} \doteq \{i \in {J}\}$ ($i$ is an inlier), compute its posterior probability given all the data up to the current time, $P[{\cal I} | y^t]$, and compare it with the alternate $P[\bar{\cal I} | y^t]$ where ${\bar {\cal I}} \doteq \{ i \notin J\}$ using the posterior ratio
\begin{equation}\label{posterior_final}
L(i | y^t) \doteq \frac{P[{\cal I} | y^t]}{P[\bar {\cal I} | y^t]}   =  \frac{p_{in}(y_i^t | y_{-i}^t)}{p_{out}(y_i^t)}\frac{\epsilon}{1-\epsilon}
\end{equation}
where  $y_{-i} \doteq \{  y_j \ | \ j \neq i\}$ are all data points {\em but} the $i$-th, $p_{in}(y_j) \doteq p(y_j \ | \ j \in J)$ is the inlier density, $p_{out}(y_j) \doteq p(y_j \ | \ j \notin J)$ is the outlier density, and  $\epsilon \doteq P(i \notin {J})$ is the prior. Note that the decision on whether $i$ is an inlier cannot be made by measuring $y_i^t$ alone, but depends on all other data points $y_{-i}^t$ as well. Such a dependency is mediated by a hidden variable, the state $x$, as we describe next.

\subsection{Filtering-based computation}

The probabilities $p_{in}(y_{{J_s}}^t)$ for any subset of the inlier set $y_{J_s} \doteq \{ y_j \ | \ j \in J_s \subset J\}$ can be computed recursively at each $t$ (we omit the subscript $J_s$ for simplicity): 
 \be
 p_{in}(y^t) = \prod_{k=1}^t p(y(k)  |y^{k-1} ).
\label{eq-recursive}
 \ee
The smoothing state $x^t$ for \eqref{eq-simpl} has the property of making ``future'' inlier measurements $y_i(t+1), i \in J$ conditionally independent of their ``past'' $y_i^{t}$: 
$ y_i (t+1) \ \perp \ y_i^t \ | \ x(t) \ \forall \ i \in J
$
as well as making time series of (inlier) data points independent of each other:  
$y_i^t \perp y_j^t \ | \ x^t \ \forall \ i \neq j \in J.
$
Using these independence conditions, the factors in \eqref{eq-recursive} can be computed via standard filtering techniques \cite{jazwinski70} 
\begin{small}
\be
p(y(k) | y^{k-1}) = \int p(y(k) | x_k) dP(x_k | x_{k-1}) dP(x_{k-1} | y^{k-1})
\label{eq-filtering}
\ee
\end{small}
starting from $p(y_J(1) | \emptyset)$, where the density $p(x_k | y^{k})$ is
maintained by a filter (in particular, a Kalman filter when all the densities at
play are Gaussian).
  Conditioned on a hypothesized  inlier set $J_{-i}$ (not containing $i$), the discriminant  $L(i | y^t, J_{-i}) =  \frac{ p_{in}(y_i^t |y_{J_{-i}}^t) } {p_{out}(y_i^t)} \frac{\epsilon}{(1-\epsilon)}$  can then be written as 
\be
{
L(i | y^t, J_{-i}) 
= \frac{ \int p_{in}(y_i^t | x^t) dP(x^t | y_{J_{-i}}^t) } {p_{out}(y_i^t) } \frac{\epsilon}{(1-\epsilon)}
}\label{posterior_approx}
\ee
   The smoothing density $p(x^t | y_{J_{-i}}^t)$ in \eqref{posterior_approx} is
maintained by a smoother \cite{andersonM79}, or equivalently a filter
constructed on the delay-line \cite{moore1973fixed}.
The challenge in using this expression is that we do not know the inlier set $J_{-i}$; to compute the discriminant \eqref{posterior_final} let us observe that 
\begin{multline}
p_{in}(y_i^t | y_{-i}^t) =  \sum_{J_{-i} \in {\cal P}_{-i}^N} p(y_i^t, J_{-i} \cup \{i\} | y_{-i}^t)   =  \\
\sum_{J_{-i} \in {\cal P}_{-i}^N} p_{in}(y_i^t|  y_{J_{-i}}^t) P[J_{-i} | y_{-i}^t] 
\end{multline}
 where ${\cal P}_{-i}^N$ is the power set of $[1, \dots, N]$ not including $i$. Therefore, 
  to compute the posterior ratio \eqref{posterior_final}, we have to marginalize $J_{-i}$, i.e., average \eqref{posterior_approx}  over all possible  $J_{-i} \in {\cal P}_{-i}^N$ 
  \be \label{BD_averaged}
L(i | y^t) = \sum_{J_{-i} \in {\cal P}_{-i}^N} L(i | y^t, J_{-i})  P[J_{-i} | y^t]
\ee 

\subsection{Complexity of the hypothesis set}

For the filtering $p(x_t | y_J^t)$ or smoothing densities $p(x^t | y_J^t)$ to be non-degenerate, the underlying model has to be {\em observable} \cite{hermannK77}, which depends on the number of (inlier) measurements $|J|$, with $|J|$ the cardinality of $J$. We indicate with $\kappa$ the minimum number of measurements necessary to guarantee observability of the model. Computing the discriminant \eqref{posterior_approx} on a {\em sub-minimal} set (a set $J_s$ with $| J_s | < \kappa$) does not guarantee outlier detection, even if $J_s$ is ``pure'' (only includes inliers). Vice-versa, there is diminishing return in computing the discriminant \eqref{posterior_approx} on a {\em super-minimal} set (a set $J_s$ with $| J_s | \gg \kappa$).
The ``sweet spot'' is a putative inlier (sub)set $J_s$, with $|J_s| \ge \kappa$, that is {\em sufficiently informative}, in the sense that the filtering, or smoothing, densities satisfy 
\be
dP(x^t | y^t_{J_s}) \simeq dP(x^t | y^t_J).
\label{eq-state-approx}
\ee
In this case, \eqref{posterior_final} which can be written as in \eqref{BD_averaged} by marginalizing over the power set not including $i$, 

can be broken down into the sum over pure ($J_{-i} \subseteq J$)  and non-pure sets ($J_{-i} \not\subseteq J$), with the latter gathering small probability\footnote{$P[J_{-i} | y_{-i}^t] $ should be small when $J_{-i}$ contains outliers, i.e. $J_{-i} \not\subseteq J$.}
\be
L(i | y^t) \simeq \sum_{J_{-i} \in {\cal P}_{-i}, \; J_{-i} \subseteq J} L(i | y^t; J_{-i}) P[J_{-i} | y_{-i}^t]
\ee
and the sum over sub-minimal sets further isolated and neglected, so
\be
L(i | y^t) \simeq \sum_{J_{-i} \in {\cal P}_{-i},\, J_{-i} \subseteq J, \, \ | J_{-i}| \ge \kappa} L(i | y^t; J_{-i}) P[J_{-i} | y_{-i}^t]. 
\ee
Now, the first term in the sum is approximately constant by virtue of   \eqref{posterior_approx} and (\ref{eq-state-approx}), and the sum $\sum P[J_{-i} | y_{-i}^t]$ is a constant. Therefore, the decision using \eqref{posterior_final} can be approximated with the decision based on \eqref{posterior_approx} up to a constant factor:
\begin{small}
\be
L(i | y^t) \simeq L(i | y^t; J_s) \sum_{\tiny \begin{array}{c} J_{-i} \in {\cal P}_{-i},\\ J_{-i} \subseteq J,\\ \ | J_{-i}| \ge \kappa\end{array}}  P[J_{-i} | y_{-i}^t] \propto  L(i | y^t; J_{s})
\ee
\end{small}
where $J_s$ is a fixed {\em pure} ($J_s  \subseteq J$) and {\em minimal} ($|J_s|=\kappa$) estimated inlier set, and the discriminant therefore becomes
\be
{
L(i | y^t; J_s) =  \frac{ \int p_{in}(y_i^t | x^t) dP(x^t | y_{J_s}^t) } {p_{out}(y_i^t) } \frac{\epsilon}{(1-\epsilon)}
}
\label{posterior_approx2}
\ee
While the fact that the constant is unknown makes the approximation somewhat unprincipled, the derivation above shows under what (sufficiently informative) conditions one can avoid the costly marginalization and compute the discriminant on any {\em minimal pure set} $J_s$. Furthermore, the constant can be chosen by empirical cross-validation along with the (equally arbitrary) prior coefficient $\epsilon$. 

Two constructive procedures for selecting a minimal pure set are discussed next.

\subsubsection{Bootstrapping} 
\label{sect-mahalanobis}

The outlier test for a datum $i$, given a pure set $J_s$, consists of evaluating \eqref{posterior_approx2} and comparing it to a threshold. This suggests a bootstrapping procedure, starting from any {\em minimal set} or ``seed'' $J_\kappa$ with $|J_\kappa| = \kappa$, by defining
\be
{\cal J}_\kappa \doteq  \{ i \ | \ L(i | y_{k_i}^t, J_\kappa) \ge \theta > 1 \}
\ee
and adding it to the inlier set:
\be
\hat J = J_\kappa \cup {\cal J}_\kappa
\ee
Note that in some cases, such as VINS, it may be possible to run this bootstrapping procedure with fewer points than the minimum, and in particular $\kappa = 0$, as inertial measurements provide an approximate (open-loop) state estimate that is subject to slow drift, but with no outliers. Note, however, that once an outlier corrupts the inlier set, it will spoil all decisions thereafter, so acceptance decisions should be made conservatively. The bootstrapping approach described above, starting with $\kappa=0$ and restricted to a filtering (as opposed to smoothing) setting, has been dubbed ``zero-point RANSAC.'' 
In particular, when the filtering or smoothing density is approximated with a Gaussian  $\hat p(x^t | y_{J_s}^t) = {\cal N}({\hat x}^t; P(t))$ for a given inlier set $J_s$, it is possible to construct the (approximate) discriminant \eqref{posterior_approx2}, or to simply compare the numerator to a threshold
\begin{small}
\begin{multline}
\int p_{in}(y_i^t | x^t) \hat p(x^t | y_{J_s}^t)dx^t \simeq {\cal G}(y_i^t - h(\hat x^t); C P(t) C^T + R) \\ \ge \frac{1-\epsilon}{\epsilon}p_{out}(y_i^t) \simeq \theta \nonumber
\end{multline}
\end{small}
where $C$ is the Jacobian of $h$ at $\hat x^t$. Under the Gaussian approximation, the inlier test reduces to a gating of the weighted (Mahalanobis) norm of the smoothing residual:
\be
i \in J \Leftrightarrow \| y_i^t - h(\hat x^t) \|_{C P(t) C^T + R} \le \tilde \theta
\label{eq-maha-gate}
\ee
assuming that $\hat x$ and $P$ are inferred using a pure inlier set that does not contain $i$. Here $\tilde \theta$ is a threshold that lumps the effects of the priors and constant factor in the discriminant, and is determined by empirical cross-validation. In reality, in VINS one must contend with an unknown parameter for each datum, and the asynchronous births and deaths of the data, which we address in Sections. \ref{sect-structure} and \ref{sect-implementation}. 

\subsubsection{Cross-validation} 
 
Instead of considering a single seed ${J}_\kappa$ in hope that it will contain no outliers, one can {\em sample} a number of putative choices $\{J_1, \dots, J_l\}$ and {\em validate} them by the number of inliers each induces. In other words, the ``value'' of a putative (minimal) inlier set $J_l$ is measured by the number of inliers it induces: 
\be
V_l = |{\cal J}_l|
\ee
and the hypothesis gathering the most votes is selected 
\be
\hat {J}= {\cal J}_{\arg\max_l(V_l)}
\ee
As a special case, when ${J}_i = \{i \}$ this corresponds to ``leave-all-out''
cross-validation, and has been called ``one-point Ransac'' in
\cite{civera20121}. For this procedure to work,
certain conditions have to be satisfied. Specifically,
\be
C_j P_{t+1|t} C_i^T \neq 0.
\label{eq-zero-ransac-cond}
\ee
Note, however, that when $C_i$ is the restriction of the Jacobian  with respect to a particular state, as is the case in VINS, there is no guarantee that the condition \eqref{eq-zero-ransac-cond} is satisfied.
{

\subsubsection{Ljung-Box whiteness test}

The assumptions on the data formation model imply that inliers are {\em conditionally} independent {\em given} the state $x^t$, but otherwise exhibit non-trivial correlations. Such conditional independence implies that the history of the prediction residual (innovation)  $\epsilon_i^t \doteq y_i^t - \hat y_i^t$ is {\em white}, which can be tested from a sufficiently long sample  \cite{ljung1978measure}. Unfortunately, in our case the lifetime of each feature is in the order of few tens, so we cannot invoke asymptotic results. Nevertheless, in addition to testing the temporal mean of $\epsilon_i^t$ and its zero-lag covariance \eqref{eq-maha-gate}, we can also test the one-lag, two-lag, up to a fraction of $k$-lag covariance. The sum of their square corresponds to a small sample version of Ljung-Box test \cite{ljung1978measure}. 

\subsection{Dealing with nuisance parameters}
\label{sect-structure}

The density $p(y^t_i | x(t))$ or $p(y^i_t | x^t)$, which is needed to compute the discriminant, may require knowledge of parameters, for instance $p_i$ in VINS \eqref{eq-fg}. 

The parameter can be included in the state, as done in \eqref{eq-fg}, in which case the considerations above apply to the augmented state $\{x, p\}$. Otherwise, if a prior is available, $dP(p_i)$, it can be {\em marginalized} via 
\be
p(y^t_i | x^t) = \int p(y^t_i | x^t, p_i)dP(p_i)
\ee
This is usually intractable if there is a large number of data points.  Alternatively, the parameter can be ``max-outed'' from the density 
\be
\hat p(y^t_i | x^t) \doteq \max_{p_i} p(y^t_i | x^t, p_i).
\ee
or equivalently $p(y^t_i | x^t, \hat p_i)$ where $\hat p_i = \arg\max_d p(y^t_i | x^t, d)$. The latter is favored in our implementation (Sect. \ref{sect-implementation}), in line with standard likelihood ratio  tests  for composite hypotheses. 
}
\section{Implementation} 
\label{sect-implementation}

The state of the models \eqref{eq-fg} and \eqref{eq-smoothing} is represented in
local coordinates, whereby $R$ and $R_{cb}$ are replaced by $\Omega ,
\Omega_{cb} \in \real^3$ such that $R = \exp(\widehat \Omega)$ and $R_{cb} =
\exp(\widehat \Omega_{cb})$. Points $p_j$ are represented in the reference frame
where they first appear $t_j$, by the triplet $\{g(t_j), y_j, \rho_j\}$ via $p_j
\doteq g(t_j) \bar y_j \exp( \rho_j)$, and also assumed constant (rigid). The
advantage of this representation is that it enables enforcing positive depth $Z
= \exp(\rho_j)$, known uncertainty of $y_j$ (initialized by the measurement
$y_j(t_j)$ up to the covariance of the noise), and known uncertainty of $g(t_j)$
(initialized by the state estimate up to the covariance maintained by the
filter). Note also that the representation is redundant, for $p_j = {g(t_j)\bar
g}{\bar g^{-1} \bar y_j} \exp(\rho_j) \doteq \tilde g(t_j) \tilde y_j
\exp(\tilde \rho_j)$ for any $\bar g\in SE(3)$, and therefore we can assume
without loss of generality that $g(t_j)$ is {\em fixed} at the current estimate
of the state, {\em with no uncertainty}. Any error in the estimate of $g(t_j)$,
say $\bar g$, will be transferred to an error in the estimate of $\tilde y_j$
and $\tilde \rho_j$ \cite{hernandezS13}.

The groups will be defined up to an arbitrary reference frame $(\bar R_i, \bar T_i)$, except for the reference group where that transformation is fixed. Note that, as the reference group ``switches'' (when points in the reference group become occluded or otherwise disappear due to tracker failure), a small error in pose is accumulated. This error affects the gauge transformation, not the {\em state} of the system, and therefore is not reflected in the innovation, nor in the covariance of the state estimate, that remains bounded. This is unlike \cite{mourikisR07}, where the covariance of the translation state $T_B$ and the rotation about gravity $\theta$ grows unbounded over time, possibly affecting the numerical aspects of the implementation.

Given that the power of the outlier test \eqref{posterior_approx2} increases with the observation window, it is advantageous to make the latter as long as possible, that is from birth to death. The test can be run at death, and if a point is deemed an inlier, it can be used (once) to perform an update, or else discarded. In this case, the unknown parameter $p_i$ must be eliminated using one of the methods described above. This is called an ``out-of-state update'' because the index $i$ is never represented in the state; instead, the datum $y_i$ is just used to update the state $x$. This is the approach advocated by \cite{mourikisR07}, and also \cite{soattoP98PAMI1,soattoP98PAMI2} where all updates were out-of-state. Unfortunately, this approach does not produce consistent scale estimates, which is why at least some of the $d_j$ must be included in the state \cite{chiusoFJS02}.

If a minimum observation interval is chosen, points that are accepted as inliers (and still survive) can be included in the state by augmenting it with the unknown parameter $p_i$ with a trivial dynamic $\dot p_i = 0$. Their posterior density is then updated together with that of $x(t)$, as customary. These are called ``in-state'' points. The latter approach is preferable in its treatment of the unknown parameter $p_i$, as it estimates a joint posterior given all available measurements, whereas the out-of-state update depends critically on the approach chosen to deal with the unknown depth, or its approximation. However, computational considerations, as well as the ability to defer the decision on which data are inliers and which outliers as long as possible, may induce a designer to perform out-of-state updates at least for some of the available measurements \cite{mourikisR07}.

The prediction for the model \eqref{eq-smoothing} proceeds in a standard manner by numerical integration of the continuous-time component. We indicate the mean $\hat x_{t | \tau} \doteq {\mathbb E}(x(t) | y^\tau)$, where $y^\tau$ denotes all available measurements up to time $\tau$; then  we have 
\be\label{state-prediction}
\begin{cases}
\hat x_{t+dt|t} = \int_{t}^{t+dt} f(x_\tau) + c(x_\tau)u(\tau) d\tau, ~~~~ x_t = \hat x_{t|t} \cr
\hat {\x}^k_{t+dt|t} = F \hat \x^k_{t|t} + C\hat x_{t|t}
\end{cases}
\ee
whereas the prediction of the covariance is standard from the Kalman filter/smoother of the linearized model. 
The update requires special attention since point features can appear and disappear at any instant. For each point $p_j$, at time 
$t + dt$ the following cases arise (in addition to the baseline models that test the instantaneous innovation with either zero-point ($m1$), or one-point RANSAC ($m2$)): \\
\noindent (i)   $ t +dt = t_j$ (feature appears): $\hat y_j \doteq y_j(t_j) \simeq y_j$ is stored and $g(t_j)$ is fixed at the current pose estimate (the first two components of $\hat x_{t+dt | t}$). \\
 \noindent (ii) $ t-k dt < t_j < t + dt$ (measurement stack is built): $y_j(t)$ is stored in ${\y}^k_j(t)$. \\
\noindent  (iii) $ t = t_j + k dt$ (parameter estimation): The measurement stack and the smoother state ${\hat \x}_{t|t_j}$ are used to infer $\hat p_j$:
\be
\hat p_j = \arg\min_{p_j}  \| {\epsilon}(t,p_j) \|
\ee
where
\be
{\epsilon}(t, p_j) \doteq {\y}_j(t) - h^k(\hat \x_{t | t_j}, p_j).
\ee
(Inlier test): the ``pseudo-innovation'' ${\epsilon}(t, \hat p_j)$ is computed and used to test for consistency with the model according to \eqref{eq-maha-gate} and, if $p_j$ is deemed an inlier: \\
\noindent (Update): the state at $t = t_j + k dt$ is computed as:
\be
\ba{c}
\hat x\\
\hat \x^k\\
\ea_{t|t} = 
\ba{c}
\hat x\\
\hat \x^k\\
\ea_{t|t_j} + L(t) {\epsilon}(t, \hat p_j) 
\ee
where $L(t)$ is the Kalman gain computed from the linearization. Alternatively,  if resources allow, we can insert $p_j$ into the state, initialized with ${\hat p}_{j_{t|t_j}} \doteq \hat p^j$ and compute the ``in-state update'':
\be
\ba{c}
\hat x\\
\hat \x^k\\
\hat p_j
\ea_{t|t} = 
\ba{c}
\hat x\\
\hat \x^k\\
\hat p_j 
\ea_{t|t_j} + L(t) {\epsilon}(t, {\hat p}_{j_{t|t_j}}) 
\label{eq-update-batch}
\ee
(iv) $t >  t_j  + k dt$: If the feature is still visible and in the state, it continues being updated  and subjected to the inlier test. This can be performed in two ways: \\
\noindent (a) -- batch update: The measurement stack $\y_j(t)$ is maintained, and the
update is processed in non-overlapping batches (stacks) at intervals $k dt$,
using the same update \eqref{eq-update-batch}, either with zero-point ($m5$) or
1-point RANSAC ($m6$)
\begin{multline}
\ba{c}
\hat x\\
\hat \x^k\\
\hat p_j
\ea_{t+kdt|t+kdt} = 
\ba{c}
\hat x\\
\hat \x^k\\
\hat p_j 
\ea_{t+kdt|t} +\\ +  L(t+kdt) {\epsilon}(t+kdt, {\hat p}_{j_{t+kdt|t}}) 
\end{multline}
after a standard robustness test on the smoothing innovation $\epsilon$; alternatively, \\
\noindent (b) -- history-of-innovation test update: The (individual) measurement $y_j(t)$ is processed {\em at each instant} while the stack $\y_j(t+dt)$ is used to test the inlier status either with zero-point ($m3$) or 1-point RANSAC ($m4$):
\begin{multline}
\ba{c}
\hat x\\
\hat \x^k\\
\hat p_j
\ea_{t+dt|t+dt} = 
\ba{c}
\hat x\\
\hat \x^k\\
\hat p_j 
\ea_{t+dt|t} + \\ + L(t+dt) \Big(y_j(t+dt) - h(\hat x_{t+dt|t}, \hat p_{j_{t+dt | t})}\Big)
\end{multline}
only for those points $j$ for which the history of the (pseudo)-innovation ${\epsilon}(t+dt, {\hat p}_{j_{t+dt |t}})$ is {\em sufficiently white,} measured as described in Sec. \ref{sect-filtering}.

Note that in the first case one cannot perform an update at each time instant, as the noise $\n_j(t)$ is not temporally white. In the second case, the history of the innovation is {\em not} used for the filter update, but just for the inlier test. Both approaches differ from standard robust filtering that only relies on the (instantaneous) innovation, without exploiting the time history of the measurements.

\section{Empirical validation}
\label{sect-experiments}

\begin{figure}[htb]
\begin{center}
\includegraphics[trim = 3cm 2.25cm 3cm
0.49cm,clip=true,width=0.49\textwidth]{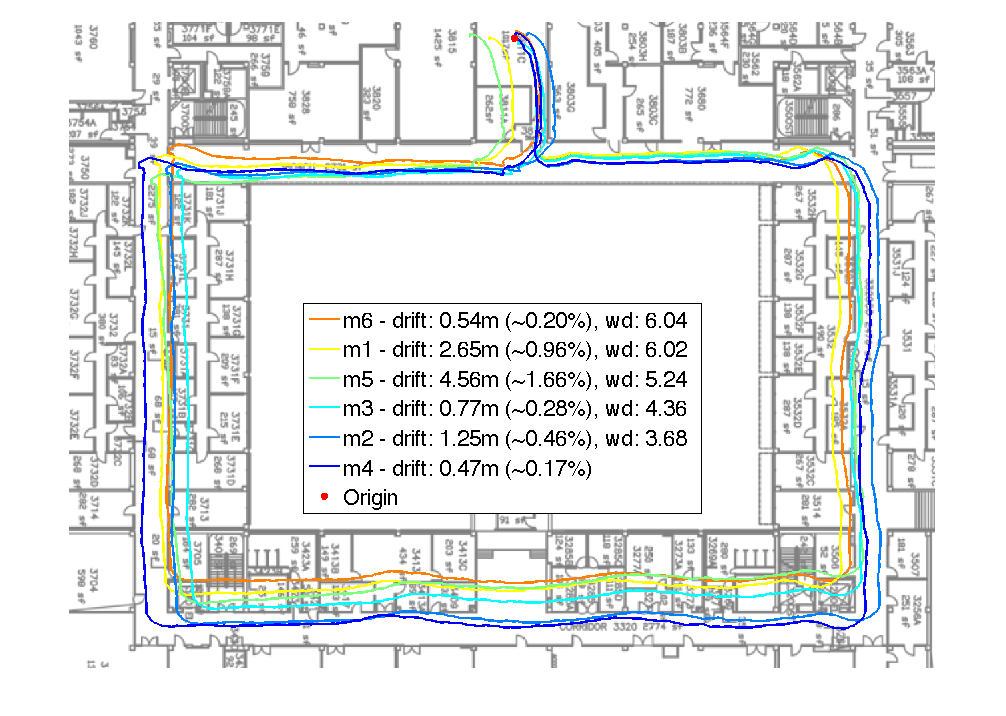}
\end{center}
\vspace{-.5cm}
\caption{\sl \small $\sim$275m loop through specular hallways with handheld motion. Note that less effective robustness strategies lead to inconsistent estimates not necessarily evident in the end point drift, with $wd$ providing a more effective ordering.}
\label{fig-specular}
\end{figure}

\begin{figure}[htb]
\begin{center}
\includegraphics[trim = 1cm 0.5cm 1cm 1cm,clip=true,width=0.45\textwidth]{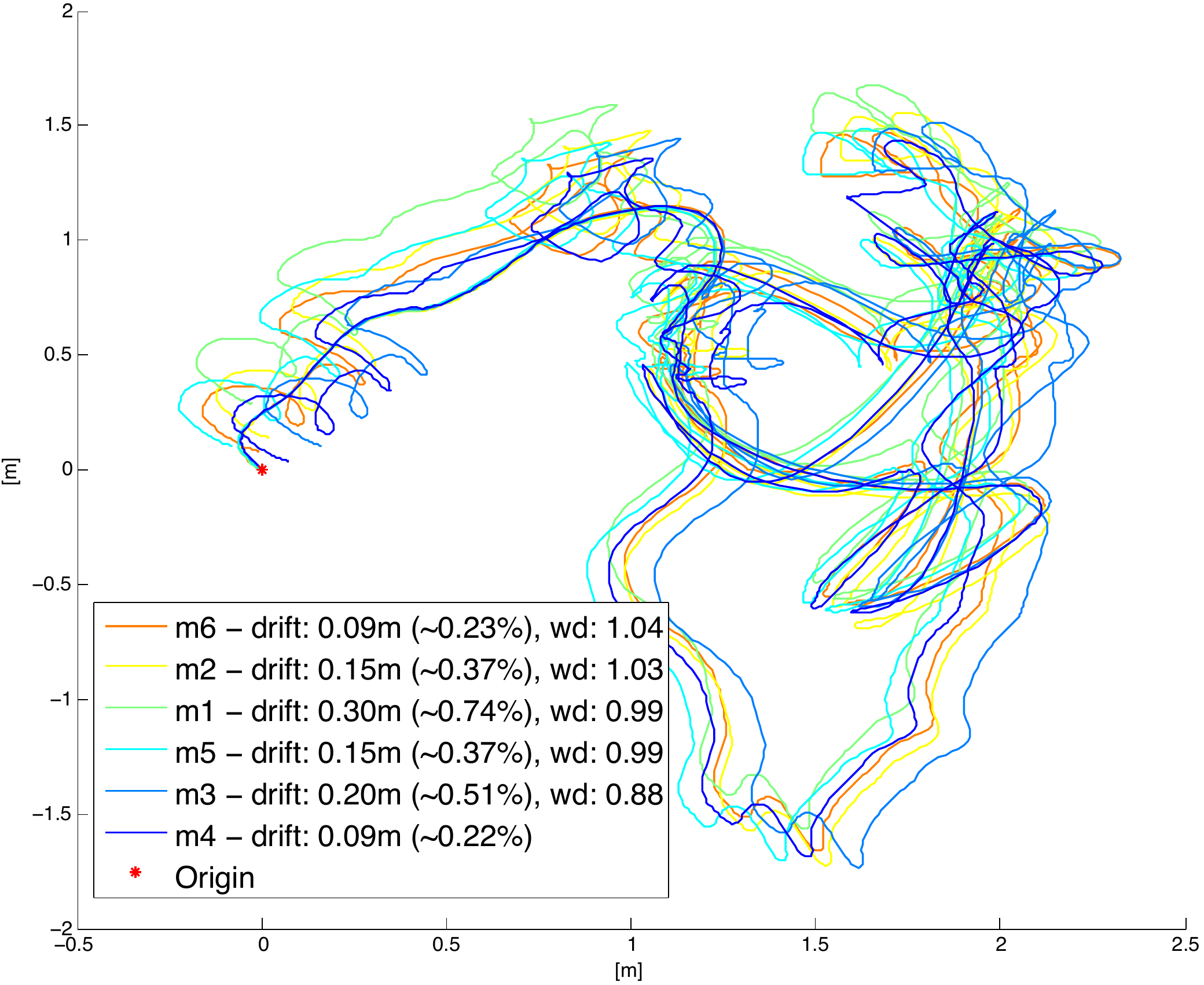}
\end{center}
\vspace{-.5cm}
\caption{\sl \small Top-down view of $\sim$40m aggressive hand-held motion loop
in a controlled laboratory environment. In the absence of significant tracking
outliers, all robustness models perform comparably, and are robust to
challenging motions/motion blur. }
\label{fig-aggressive}
\end{figure}

\begin{figure}[htb]
\begin{center}
\includegraphics[trim = 2.25cm 2.25cm 2.3cm
1.5cm,clip=true,width=0.49\textwidth]{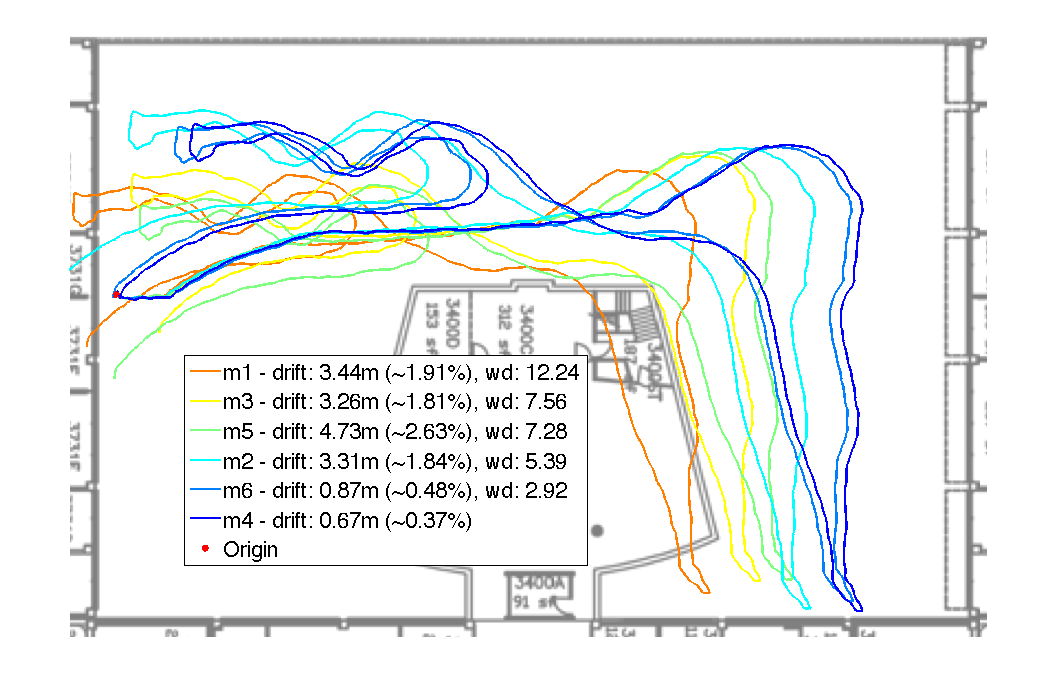}
\end{center}
\vspace{-.5cm}
\caption{\sl \small $\sim$180m loop through natural forested area with abundant
occlusions due to foliage. Poor outlier handling in natural environments
severely degrades performance.}
\label{fig-occlusions}
\end{figure}

\begin{figure}[htb]
\begin{center}
\includegraphics[trim = 3.5cm 4cm 4cm
4cm,clip=true,width=0.49\textwidth]{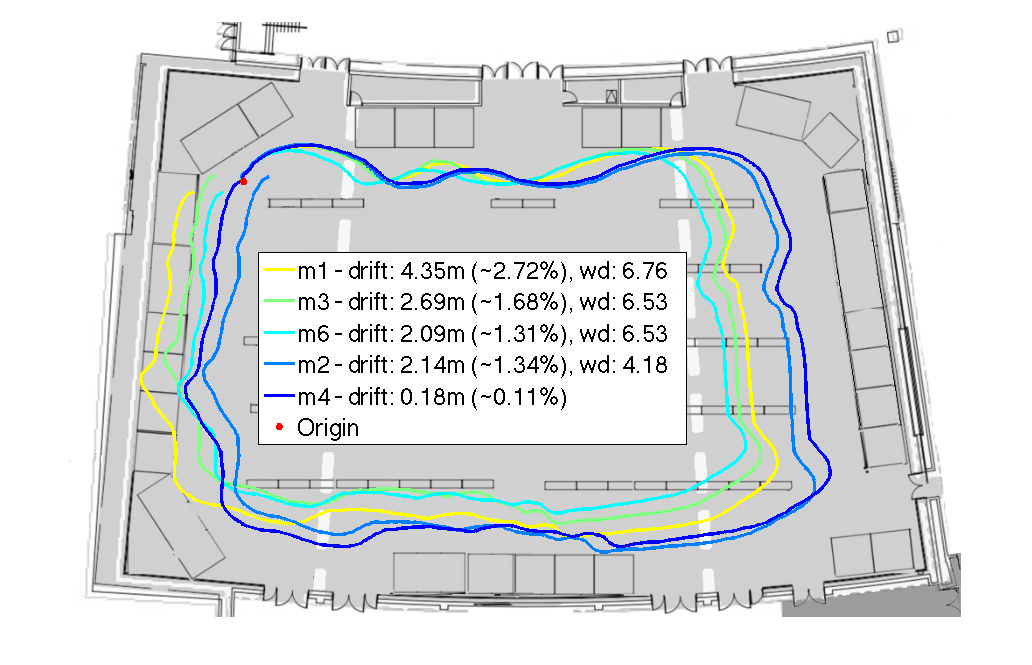}
\end{center}
\vspace{-.5cm}
\caption{\sl \small $\sim$160m loop through a crowded hall during a poster session, with many independently moving objects constantly in view. Less
effective robustness strategies see similar estimate biases as they do for
occlusions and specularities. 
}
\label{fig-movingobj}
\end{figure}

To validate our analysis and investigate the design choices it suggests, we
report quantitative comparison of various robust inference schemes on real data
collected from a hand-held platform in artificial, natural, and outdoor
environments, including aggressive maneuvers, specularities, occlusions, and
independently moving objects. Since no public benchmark is available, we do not
have a direct way of comparing with other VINS systems: We pick a
state-of-the-art evolution of \cite{jonesS09}, already vetted on long driving
sequences, and modify the outlier rejection mechanism as follows: $(m1)$
Zero-point RANSAC; $(m2)$ same with added 1-point RANSAC, ; $(m3)$ $m1$ with
added test on the history of the innovation; $(m4)$ same with 1-point RANSAC;
$(m5)$ $m3$ with zero-point RANSAC and batch updates; $(m6)$ same with 1-point
RANSAC. We report end-point open-loop error, a customary performance measure,
and trajectory error, measured by dynamic time-warping distance $wd$, relative
to the lowest closed-loop drift trial. Figures
\ref{fig-specular} to \ref{fig-movingobj} show a comparison of the six schemes
and their ranking. All trials use the same settings and tuning, and run at
frame-rate on a 2.8GHz Core i7 processor, with a 30Hz global shutter camera and
an XSense MTi IMU. The upshot is that the most effective strategy is a whiteness
testing on the history of the innovation in conjunction with 1-point RANSAC.
Based on $wd$, the next-best method ($m2$, without the history of the
innovation) exhibits a performance gap equal to the gap from it to the
last-performing.

\section{Discussion}

We have described several approximations to a robust filter for visual-inertial
sensor fusion (VINS) derived from the optimal discriminant, which is
intractable. This addresses the preponderance of outlier measurements typically provided by a visual
tracker, Sect. \ref{sect-filtering}. Based on modeling considerations, we have
selected several approximations, described in Sect. \ref{sect-implementation},
and evaluated them in Sect. \ref{sect-experiments}.

Compared to ``loose integration'' systems
\cite{weiss2013monocular,engel14ras,weiss2011monocular} where pose estimates
are
computed independently from each sensory modality and fused {\em post-mortem},
our approach has the advantage of remaining within a bounded set of the true
state trajectory, which cannot be guaranteed by loose integration
\cite{hernandezS13}. Also, such systems rely on vision-based inference to
converge to a pose estimate, which is delicate in the absence of inertial
measurements that help disambiguate local extrema and initialize pose estimates.
As a result, loose integration systems typically require careful initialization
with controlled motions.

Motivated by the derivation of the robustness test, whose power increases with
the window of observation, we adopt a smoother, implemented as a filter on the
delay-line \cite{andersonM79}, like \cite{mourikisR07,liMourikis13}. However,
unlike the latter, we do not manipulate the measurement equation to remove or
reduce the dependency of the (linearized approximation) on pose parameters.
Instead, we either estimate them as part of the state if they pass the test, as
in \cite{jonesS09}, or we infer them out-of-state using maximum likelihood, as
standard in composite hypothesis testing. 

We have tested different options for outlier detection, including using the
history of the innovation for the robustness test while performing the
measurement update at each instant, or performing both simultaneously at
discrete intervals so as to avoid overlapping batches. Our experimental
evaluation has shown that in practice the scheme that best enables robust pose
and structure estimation is to perform instantaneous updates using 1-point
RANSAC and to continually perform inlier testing on the history of the
innovation.

\begin{footnotesize}
\bibliographystyle{IEEEtran}
\bibliography{extra,total2,filter,self}
\end{footnotesize}

\end{document}